\documentclass[runningheads]{llncs}
\usepackage{graphicx}
\usepackage{subfigure}
\usepackage{amsmath}
\usepackage{algorithm}
\usepackage{threeparttable}
\usepackage{multicol}
\usepackage{multirow}
\usepackage{booktabs}
\usepackage{subfigure}
\usepackage{epstopdf}
\usepackage{mathptmx}
\usepackage{float}
\usepackage{algorithmicx}
\usepackage{algpseudocode}
\usepackage{graphicx}
\usepackage{amsmath}
\usepackage{threeparttable}
\usepackage{subfigure}
\usepackage{mathptmx}
\usepackage{url}
\usepackage{hyperref}
\usepackage{comment}
\usepackage{subfigure}
\usepackage{float}
\usepackage[misc]{ifsym}

\begin{document}
\title{SECaps: A Sequence Enhanced Capsule Model for Charge Prediction}%\thanks{Supported by organization x.}}

\author{Congqing He\inst{1} \and
Li Peng\inst{1}\textsuperscript{(\Letter)} \and
Yuquan Le \inst{1} \and
Jiawei He\inst{1} \and
Xiangyu Zhu\inst{2}}
\authorrunning{C. He et al.}
\institute{ \Letter  Li Peng\at}

\institute{College of Computer Science and Electroic Engineering, Hunan University, China \and
 JD Digits, China\\
\email{\{hecongqing,rj\_lpeng,leyuquan,hejiawei\}@hnu.edu.cn\\ zhuxiangyu3@jd.com}
}

%\institute{ \Letter  Li Peng\at}

%
\maketitle              % typeset the header of the contribution
\begin{abstract}
Automatic charge prediction aims to predict appropriate final charges according to the fact descriptions for a given criminal case.
Automatic charge prediction plays a critical role in assisting judges and lawyers to improve the efficiency of legal decisions, and thus has received much attention. Nevertheless, most existing works on automatic charge prediction perform adequately on high-frequency charges but are not yet capable of predicting few-shot charges with limited cases.
%On the other hand, some works have shown the benefits of capsule network, which is a powerful technique.
In this paper, we propose a \textbf{S}equence \textbf{E}nhanced \textbf{Caps}ule model, dubbed as SECaps model, to relieve this problem.
%This motivates us to propose a \textbf{S}equence \textbf{E}nhanced \textbf{Caps}ule model, dubbed as SECaps model, to relieve this problem.
%More specifically, we propose a  new basic structure, seq-caps layer, to enhance capsule by taking sequence information into account. In addition, we construct our SECaps model by making use of seq-caps layer.
Specifically, following the work of capsule networks, we propose the seq-caps layer, which considers sequence information and spatial information of legal texts simultaneously. Then we design a attention residual unit, which provides auxiliary information for charge prediction. In addition, our SECaps model introduces focal loss, which relieves the problem of imbalanced charges.
Comparing the state-of-the-art methods, our SECaps model obtains 4.5\% and 6.4\% absolutely considerable improvements under Macro F1 in Criminal-S and Criminal-L respectively. The experimental results consistently demonstrate the superiorities and competitiveness of our proposed model.

\end{abstract}

\section{Introduction}
The task of automatic charge prediction is to help lawyers or judges to determine appropriate charges (e.g., fraud, robbery or larceny) according to a given case. The automatic charge prediction plays a critical role in many legal intelligent scenarios (e.g., legal assistant systems or legal consulting). The legal assistant system can improve the efficiency of professionals. The legal consulting is benefit for people who are unfamiliar with legal terminology of their interested cases. Therefore, automatic charge prediction is an extremely beneficial topic for many legal intelligent scenarios.

Most existing works of automatic charge prediction can be divided into three categories. The first categories are usually mathematical or quantitative \cite{Kort1957Predicting,Nagel1964Applying}, which are restricted to a small dataset with few labels. The second categories use a lot of manpower to design legal text features, and then the machine learning algorithms are applied. For example, Liu et al. \cite{Liu2006Exploring} utilize word-level and phrase-level features and K-Nearest Neighbor (KNN) method to predict charges. Liu et al. \cite{Liu2015Predicting} use Support Vector Machine (SVM) for preliminary article classification, and then re-rank the results by using word level features and co-occurence tendency among articles. Katz et al. \cite{Katz2014A} extract efficient features from case profiles (e.g., dates, locations, terms, and types). However, the shallow textual features of human designs require a lot of manpower and have limited ability to capture the semantic information of legal texts. Recently, owing to the success of deep neural networks on nature language processing tasks \cite{le2018acv}, some popular neural network methods apply on automatic charge prediction task \cite{Luo2017Learning,hu2018few}, obtaining attractive performance. For example, Luo et al. \cite{Luo2017Learning} propose an attention-based neural network for charge prediction by incorporating the relevant law articles. This work is not yet capable of predicting few-shot charges with limited cases. Hu et al. \cite{hu2018few} propose attribute-attentive charge prediction model to alleviate few-shot charges problem. In the other hand, Zhao et al. \cite{Zhao2018Investigating} apply the capsule network \cite{sabour2017dynamic} to the text classification scene and  achieve attractive performance.

Inspired by the above observations, in this paper we propose a Sequence Enhanced Capsule model, dubbed as SECaps model.
The SECaps model  proposes the seq-caps layer, which can consider sequence information and spatial information of legal texts simultaneously. Then the model designs a attention residual unit, which can provide auxiliary information for charge prediction. In addition, the model introduces focal loss, which can relieve the problem of imbalanced charges.
%The SECaps model belongs to the deep neural network method, which can better capture the legal text semantic information. What's more, it can deal with the problem of the few-shot charges.

To summarize, the main contributions of this paper are:

${\bullet}$\quad %We propose a Sequence Enhanced Capsule model that not only captures the prominent features and the semantic information of legal texts in a better way, but also has a competitive performance on few-shot charges problem.
We propose a Sequence Enhanced Capsule model that not only considers sequence information and spatial information of legal texts simultaneously, but also has a competitive performance on the problem of few-shot charges.

${\bullet}$\quad Our SECaps model introduces focal loss, which first appear on object detection problems and is able to alleviate the problem of imbalanced charges to some extent.

${\bullet}$\quad Comparing the state-of-the-art methods, our SECaps model achieves 4.5\% and 6.4\%  absolutely considerable improvements under Macro F1 in Criminal-S and Criminal-L  respectively. The experimental results consistently demonstrate the superiorities and competitiveness of our proposed model.

%The rest of the paper is organized as follows. Section 2 surveys the related works. Section 3 introduces the detail descriptions of SECaps model. The performance evaluation is given in section 4. Section 5 concludes the paper.

\section{Model}
%In this section, we propose the Sequence Enhanced Capsule model, dubbed as SECaps model, for charge prediction task. We mainly focus on the few-shot charge problem in the charge prediction modeling. Our SECaps model combines wildly used deep learning method for natural language processing with a newly proposed capsule network. In what follows, we first review the capsule network, for ease of understanding our proposed SECaps model (Section 3.1). Then we introduce the basic structure of SECaps model: seq-caps layer (Section 3.2). Finally, we provide the details architecture of the proposed SECaps model (Section 3.3).

%The architectural overview of our SECaps model is introduced in this section. Section \ref{sect:capsule_network} reviews the capsule networks for understanding our model easily. Then the basic structure of SECaps model (seq-caps layer) is introduced in Section \ref{sect:seq_caps}. Finally, we describe the detailed architecture of the SECaps model in Section \ref{sect:SECaps}.

\subsection{Capsule Network}\label{sect:capsule_network}

Capsule network proposed by Sabour et al. \cite{sabour2017dynamic} has shown strong competitiveness in the field of images. %A capsule is a group of neurons whose activity vector represents the feature parameters of a specific type of entity such as an object or an object part \cite{DBLP:conf/nips/SabourFH17}. The length of the capsule represents the probability of feature existence and the orientation represents the feature parameters.
 Capsule network adopts dynamic routing mechanism, and routes the lower-level capsules to higher-level capsules.

%Capsule network which initially aims to solve object recognition task show its competitive performance in MNIST task. A capsule is a group of neurons whose activity vector represents the feature parameters of a specific type of entity such as an object or an object part \cite{DBLP:conf/nips/SabourFH17}. We can view a capsule as a vector that describe a specific features. The length of the capsule represent the probability that the feature exists and the orientation of that represent the feature parameters.

%A capsule is a basic element of a capsule network like a neuron is a basic element of a neural network. Neural network transform lower-layer neurons to higher-layer neurons by making use of affine transformation and a non-linear activate function e.g. $relu$, $sigmoid$, $tanh$. Whereas, in capsule network, a dynamic routing mechanism is used to send lower-layer capsules predictions to higher-layer capsules that agrees with the lower-layer capsules.

Define lower-level capsules as $u=(u_1, u_2, \cdots, u_n)$ and higher-level capsules as $v=(v_1, v_2, \cdots, v_m)$, where $u_i \in R^d$ is the $i$-th capsule in lower-level and $v_j \in R^p$ is the $j$-th capsule in higher-level, $d$ and $p$ represent the dimension of lower-level capsules and the dimension of higher-level capsules respectively. The dynamic routing mechanism have the follows two steps:

${\bullet}$ \textbf{Linear transformation}. In this step, an intermediate feature vector $u_{j|i}$ of $u_i$ is produced by multiplying the output $u_i$ by a weight matrix $W_{ij}$.
\begin{equation}
u_{j|i} = W_{ij} u_i.
\end{equation}
Where $W_{ij}$ is a weight matrix which connects between lower-level $u_i$ and higher-level $v_j$.
%There is a weight matrix $W_{ij}$ in each connection between lower-level $u_i$ and higher-level $v_j$. 
There are $n \times m$ weight matrices $W_{ij}$ between two capsule layers. However, the weight matrices $W_{ij}$ produces the large amount of the parameters.
%One could be worry about overfitting due to the large amount of the parameters.
In order to reduce parameters, we introduce share weight mechanism, which is similar with Zhao et al. \cite{Zhao2018Investigating}. In share weight mechanism, the connection between all the lower-level capsules and the $j$-th capsule in higher-level share a common weight matrix $W_j$, so the intermediate feature vector $u_{j|i}$ is computed as follows
\begin{equation}
u_{j|i} = W_j u_i.
\end{equation}

${\bullet}$ \textbf{Clustering for lower-level capsules}. In this step, the dynamic routing mechanism minimizes an agglomerative fuzzy k-means clustering-like loss function as follows:
\begin{equation}
\begin{array}{lr}
\min\limits_{C,S}\{loss(C,S)=-\sum_{i,j} c_{ij}\left<u_{j|i}, v_j\right> + \alpha \sum_{i,j} c_{ij}\text{log}c_{ij}\}
\\
\text{s.t. } c_{ij}>0, \sum_{j=1}^n c_{ij} = 1, \lVert v_j \rVert \leq 1
\end{array}
\end{equation}
where $C = [c_{ij}]$ is an $n$-by-$m$ partition matrix,
%$c_{ij}$ represents the association degree of membership of the intermediate feature of $i$-th lower-layer capsule $u_{j|i}$ to the $j$-th cluster $s_j$,
$c_{ij}$ represents the association degree of $i$-th lower-level capsule $u_{j|i}$ to the $j$-th cluster $s_j$,
$S = [s_1, s_2, \cdots, s_m]^T$  is $m$ cluster centers. Then, similar to Hinton et al. \cite{sabour2017dynamic}, we use a non-linear ``squashing'' function to ensure that short vectors get shrunk to almost zero length and long vectors get shrunk to a length slightly below 1, and thus get the higher-level capsule. The formula of ``squashing'' is described as follows:
\begin{equation}
\text{squash}(s_j) = \frac{{\lVert s_j \rVert}^2}{1 + {\lVert s_j \rVert}^2 } \frac{s_j}{\lVert s_j \rVert }.
\end{equation}
Deriving the coordinate descent updates of $C$ and $S$, we obtain the updates in Algorithm \ref{al:dr} \cite{optimizationDR}.
\begin{algorithm}[H]
\caption{Dynamic Routing} %Ëã·¨µÄÃû×Ö
\label{al:dr}
%\begin{flushleft}
{\bf Input: }
 $u_{j|i}$,   %/*\emph{intermediate feature vector}*/\\
 $r$ \\%/*\emph{routing iterations parameter}*/ \\
{\bf Output: }
 $v_j$ %/*\emph{the high-layer capsule}*/
%\end{flushleft}
\begin{algorithmic}[1]
\State  for all capsule $i$ in lower-level and capsule $j$ in higher-level: $b_{ij} = 0$.
\For{ $r$ iterations}
\State  for all capsule $i$ in lower-level and capsule $j$ in higher-level:
\State  \text{ }\text{ }\text{ }\text{ }$c_{ij} = \frac{exp(b_{ij})}{\sum_k exp(b_{ik})}$.
\State  for all capsule $j$ in higher-level capsule:
%\State \text{ }\text{ }\text{ }\text{ }$\hat{s}_j = \sum_{i=1}^{m} c_{ij}u_{j|i}$,
 %$s_j = \frac{\hat{s}_j}{\lVert \hat{s}_j \rVert}$
\State \text{ }\text{ }\text{ }\text{ }$s_j = \sum_{i=1}^{m} c_{ij} \cdot u_{j|i}$,
%\State  \text{ }\text{ }\text{ }\text{ }$v_j = \frac{{\lVert \hat{s}_j \rVert}^2}{1 + {\lVert \hat{s}_j \rVert}^2 }s_j$
\text{ }$v_j = \text{squash}(s_j)$,
\State \text{ }\text{ }\text{ }\text{ }$b_{ij}=b_{ij} + u_{j|i} \cdot v_j $.
\EndFor
\State \Return $v_j$
\end{algorithmic}
\end{algorithm}

%And we can see the ``squashing'' function in this dynamic routing mechanism is
%\begin{equation}
%squashing(s_j) = \frac{{\lVert \hat{s}_j \rVert}^2}{1 + {\lVert \hat{s}_j \rVert}^2 }s_j
%\end{equation}

\subsection{Seq-caps Layer}\label{sect:seq_caps}

\begin{figure}[htbp]
\begin{center}
\includegraphics[scale=0.5]{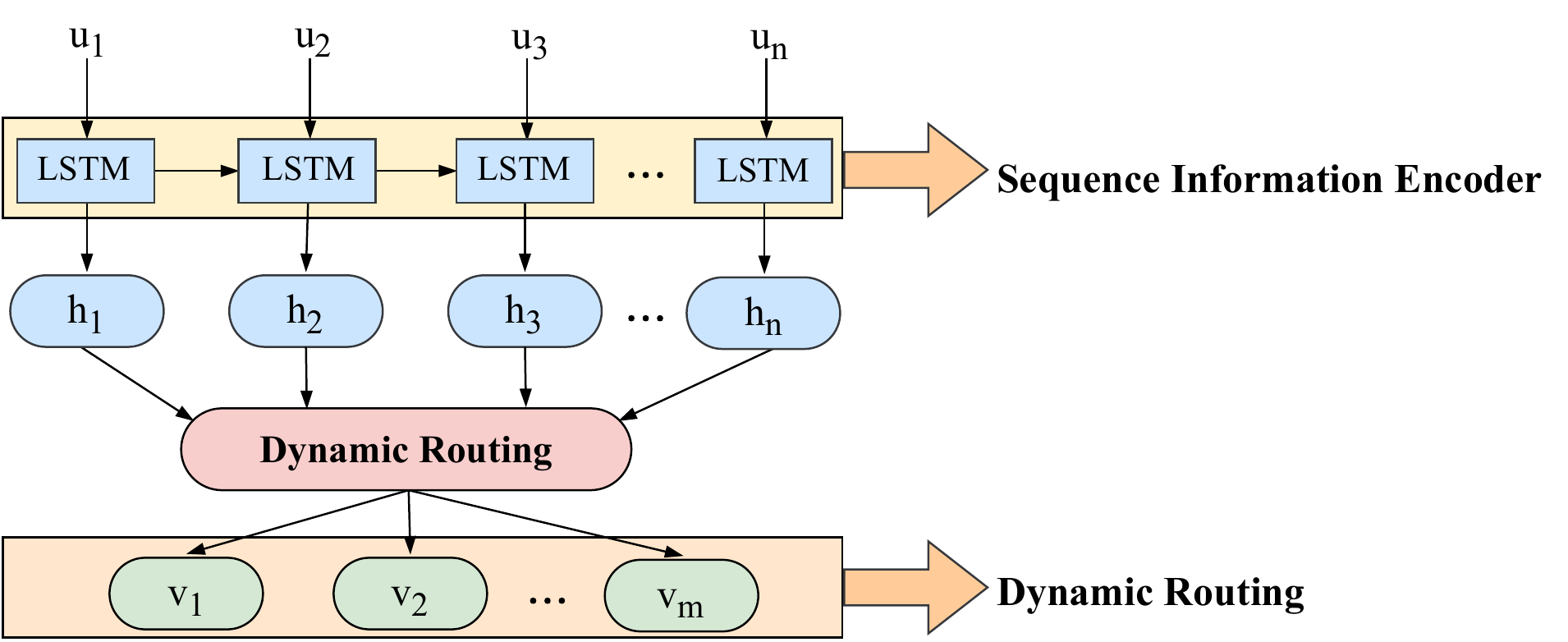}
\caption{The framework of seq-caps layer, the input capsules $u = \{u_1, u_2, \cdots, u_n\}$ is lower-level capsules and the output capsules of the seq-caps layer is $v={\{v_1, v_2, \cdots, v_m\}}.$}
\label{fig:caps}
\end{center}
\end{figure}

Capsule networks treat a feature as a activity vector, it can be used in many nature language processing (NLP) tasks~\cite{Zhao2018Investigating}~. Generally, the input of many NLP tasks is a sequence of words which represents a sentence or a text.
Each word of the sequence is often transformed into the distributed representation of word, due to the success of word embeddings~\cite{mikolov2013distributed}~.
The word distributed representation can be seen as a activity vector, and thus a sequence of words can be seen as a group of capsules. We can use capsule networks in these NLP tasks as long as we set the first layer of capsule networks to words distributed representation of words sequence.

However, the higher-level capsules capture the key information of lower-level capsules by making use of fuzzy clustering. This lead to the higher-level capsules loss the sequence information of the input word sequence. In fact, a word is often highly correlated with its context. Losing sequence information weaken the performance of capsule network in NLP tasks. Therefore, we propose a new basic structure, named seq-caps layer, to enhance the capsule layer by taking the sequence information into account.

Figure \ref{fig:caps}  shows the framework of our seq-caps layer. Suppose the input capsules of the seq-caps layer is $u = \{u_1, u_2, \cdots, u_n\}$, our seq-caps layer has the follows two component:

${\bullet}$  \textbf{Sequence Information Encoder}. It uses a Long Short-Term Memory \cite{hochreiter1997long} (LSTM) encoder as a sublayer to restore sequence information of the input capsules. In this step, we get hidden layer $h_1, h_2, \cdots, h_n = LSTM(u_1, u_2, \cdots, u_n)$.

${\bullet}$  \textbf{Dynamic Routing}. It transforms the hidden layer to higher-level capsules by using dynamic routing mechanism (Algorithm \ref{al:dr}). In this step, we get higher-level capsules $\{v_1, v_2, \cdots, v_m\}$.

\subsection{SECaps Model}\label{sect:SECaps}

\begin{figure*}[!htbp]
\begin{center}
\includegraphics[scale=0.65]{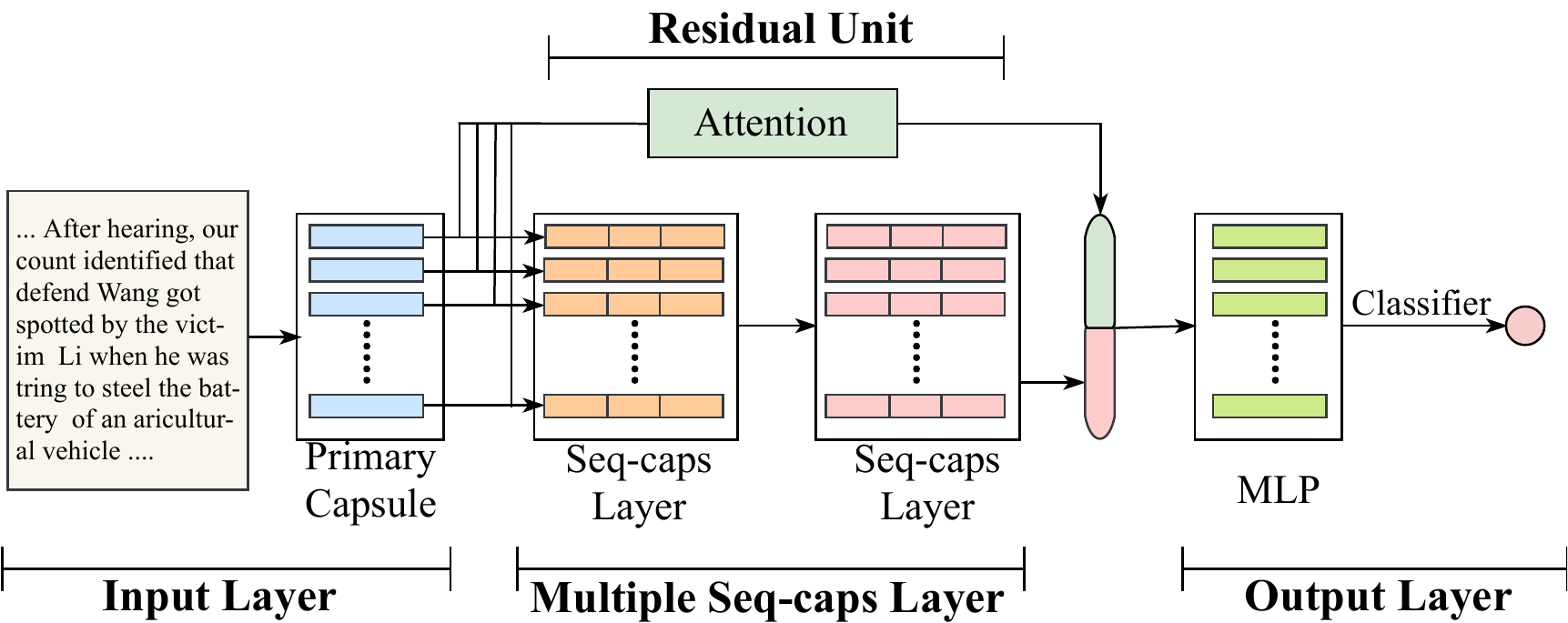}
\caption{The architecture of SECaps model, including Input layer, Multiple seq-caps layer, Attention, and Output layer.}
\label{fig:secapsnet}
\end{center}
\end{figure*}

In charge prediction task, the fact description of a case can also be seen as a sequence of words $x = \{x_1, x_2, \cdots, x_n\}$, where $n$ is length of the fact description, $x_i$ is a word. Given the fact description $x$, the charge prediction task aims to predict a charge $y \in Y$ from a charge set $Y$.

In real world, some charges (e.g., \textit{theft}, \textit{intentional injury}) have large amount of cases, while others like \textit{scalping relics}, \textit{disrupting the order of the court} just have few cases. This is so called few-shot problem. Traditional models pay much attention to charges which have large amount of cases and thus ignore these few-shot charges. In order to mitigate the effect of few-shot problem, our SECaps model combines seq-caps layer with the focal loss \cite{lin2018focal}, in which, seq-caps layer captures the prominent features and the semantic information of legal texts in a better way and focal loss is able to alleviate category imbalances to some extent.

Our SECaps model includes four parts: Input layer, Multiple seq-caps layer, Attention, Output layer. Figure \ref{fig:secapsnet} show the architecture of our SECaps model.

\textbf{Input layer:} In this part, we treat the fact description of a case as a sequence of words $x = \{x_1, x_2, \cdots, x_n\}$, then, each word of the sequence is transformed to primary capsule.

\textbf{Multiple seq-caps layer:} This part has two seq-caps layers. We treat the word embeddings as primary capsules, and then transfer primary capsules to higher-level capsules. The seq-caps layer outputs advanced semantic representation which are captured from fact description of a case. Meanwhile, seq-caps layer restores the sequence information of fact description, which is key factor for charge prediction.

\textbf{Attention:} When the multiple seq-caps layer aggregates  primary capsules into higher-level capsules, the model only focuses on the most important legal case's information. The similar as He et al.~\cite{he2016deep}~, we propose a new residual unit to improve the generalization and provide auxiliary information for charge prediction. Our model introduces attention mechanism as the residual unit, to encode the primary capsule which can capture the global context information.
%\textbf{Attention. } Attention mechanism is an integral part of compelling sequence modeling and transduction models in various tasks, allowing modeling of dependencies between input sequence and the context vector \cite{DBLP:conf/nips/2017}. Multiple seq-caps layer part can capture some prominent features of the fact description of a case, but cannot gain the global context information. To overcome this shortcoming, we introduce Attention part. The Attention part aims to encode the global context information of the fact description.
Suppose the primary capsules from the input part is ${\{t_1, t_2, \cdots, t_n\}}$, the residual unit's vector $c$ is computed as follows:

\begin{align}
&e_i = tanh(W t_i + b)\\
&\alpha_i = \frac{exp(e_i)}{\sum_j exp(e_j)}\\
&c = \sum_i \alpha_i t_i
\end{align}
where $W$ is a weight matrix and $b$ is bias.

\textbf{Output layer:} In order to consider prominent features and the global context information together, we first flatten all the feature vectors from the Multiple seq-caps layer, and concatenate with the global context vector $c$. Then, we use a fully connected network and softmax function to generate the probability $y = (y_1, y_2, \cdots, y_k)$, where $k$ is the number of charge. %This part considers both the prominent features of fact description and the global context information, make our model more robust.
As for loss function, we apply focal loss to SECaps model. Focal loss is proposed for dense object detection initially, which address the few-shot problem by reshaping the standard cross entropy loss such that it down-weights the loss assigned to well-classified examples \cite{lin2018focal}. It can be calculate as follows:
\begin{equation}
FL(y_t) = \alpha_t (1-y_t)^\gamma log(y_t)
\end{equation}
where $y_t$ is the $t$-th output of $y$, $\alpha_t \in [0, 1]$ is  weighting factor and $\gamma$ is the focusing parameter.

\section{ Experiments}

%In order to evaluate the effectiveness of our SECaps model for charge prediction, we conduct experiment on several real-world datasets and compare our model with several state-of-the-art baselines.

\subsection{Dataset and Evaluation Metrics}

%In this section, we introduce the datasets, evaluation metrics, all baselines and experimental details. At first, we introduce the datasets, and then describe the scoring formula in detail. Next, we describe the baseline methods. Finally, we present  the relevant experimental parameters in detail.

\subsubsection{Datasets}
%Same to the recent work on charges prediction \cite{hu2018few},  the proposed model is evaluated on dataset of Hu et al, which was published by the Chinese government from China Judgments Online\footnote{http://wenshu.court.gov.cn}.
We employ three public datasets\footnote{\url{https://thunlp.oss-cn-qingdao.aliyuncs.com/attribute\_charge.zip}.}~\cite{hu2018few}~ for charge prediction, which was published by the Chinese government from China Judgments Online\footnote{\url{http://wenshu.court.gov.cn}}. For each case in these datasets are constituted by several parts, such as fact, charges, and attributions. Three datasets which contain the same number of charges but the different number of scales, named as Criminal-S (small), Criminal-M (medium) and Criminal-L (large). The statistics of our datasets are reported in Table \ref{tab.datasets stat}.

%Besides, in order to examine the generalization of our method on few-shot charges, we adopt three datasets which contain the same number of charges but the different number of scales, named as Criminal-S (small), Criminal-M (medium) and Criminal-L (large). The statistics of our datasets are reported in Table \ref{tab.datasets stat}.

\begin{table}[!htbp]
\centering
\caption{The statistics of different datasets}
\label{tab.datasets stat}
\setlength{\tabcolsep}{3.5mm}{
\begin{tabular}{c|ccc}
\hline
{\bf  Datasets} & {\bf Criminal-S}& {\bf Criminal-M}& {\bf Criminal-L}\cr
\hline
 train & {61,589}&{153,521}&{306,900}\cr
 valid &{7,755} & {19,250} & {38,429}\cr
 test &{7,702} & {19,189} & {38,368} \cr
\hline
\end{tabular}
}
\end{table}

\subsubsection{Evaluation Metrics}

Following previous works on charge prediction \cite{Luo2017Learning,hu2018few}, we employ Accuracy (Acc.), Macro Precision (MP), Macro Recall (MR) and Macro F1 (MF) as our main evaluation metrics.

\subsection{Baselines}
The same as Hu et al.~\cite{hu2018few}~, we select several representative text classification models and two state-of-the-art methods for charge prediction as baselines.

TFIDF+ SVM is a simple machine learning model based on Support Vector Machine (SVM) \cite{suykens1999least} with linear kernel, extracting text features from term-frequency inverse document frequency (TFIDF) \cite{salton1988term} as input. Then two  based deep learning model are also to compare with our SECaps model, the first is Convolutional Neural Network (CNN) \cite{Kim2014Convolutional} which is to encode fact descriptions  with multiple filter widths, and the second  employs  a two-layer LSTM \cite{hochreiter1997long} with a max-pooling  layer as the  fact encoder.

Moreover, to future illustrate the effectiveness of our model, we compared our model with two latest similar tasks, Fact-Law Attention Model \cite{Luo2017Learning} and Attribute-attentive Charge Prediction Model \cite{hu2018few}. Fact-Law Attention Model is an attention-based neural network method for charge prediction
task, and  Hu et al.~\cite{hu2018few}~ propose Attribute-attentive Charge Prediction Model which can infer the attributes and charges simultaneously.

\subsection{Experiment Settings}

Since all the case documents have been  employed THULAC\footnote{\url{https://github.com/thunlp/THULAC-Python}} for word segmentation and set each document  maximum length to 500. For the TFIDF+SVM, the experiment is established by  extracting the feature size to $2,000$ and using SVM with linear kernel for training. Moreover, to make a fair comparison, we establish  a set of neural models. We employ word2vec \cite{mikolov2013distributed} for word embedding with size to 100 before the experiment. For the CNN and the LSTM, by setting  the filter widths to ${\{2,3,4,5\}}$ with each filter size to 25 for consistency and the hidden state size to 100 of LSTM  respectively. What's more, two recent models which proposed by Luo et al. \cite{Luo2017Learning} and Hu et al. \cite{hu2018few} respectively, the parameters remain the same as the original paper.

%We set word embedding to 100 and use Adam(Kingma and Ba, 2015) as the objective function where the learning rate is setting to 0.001. Set dropout rate(Srivastava et al., 2014) to 0.2 , and the batch size to 64 .

Our SECaps model uses the Adam \cite{kingma2014adam} optimization method to minimize the focal loss \cite{lin2018focal} over the training data. For hyperparameters of Adam and focal loss, we keep it consistent with the original papers since better performance in their papers. %by setting  $\alpha$=0.25 and $\gamma$ = 2 and the learning rate to 0.001 respectively. 
Our SECaps model has two seq-caps layers, and we set different hyperparameters which is shown in Table \ref{tab.SepCapsNet param}. Then our model utilizes two fully connected layers by setting to $1024 \times 512$. %We train the model for a fixed number of epochs  and monitor its performance on the validation set. Once the training is finished, we choose the model with the best accuracy score on the validation dataset as our final model and evaluate its performance on the test dataset.

%Then our model utilizes two fully connected layers by setting to $1024 \times 512$ which can enhance maps the ``distributed feature representation'' learned to the sample markup space. Additionally, we make use of the batch normalization \cite{ioffe2015batch} to reduce overfitting in the fully connected layer during training. We train the model for a fixed number of epochs  and monitor its performance on the validation set. Once the training is finished, we choose the model with the best accuracy score on the validation dataset as our final model and evaluate its performance on the test dataset.

\begin{table}[!htbp]
\centering
\caption{Hyperparameters for two seq-caps layers. CapsNums, CapsDims, Routing and  Hidden units represent the number of capsules, the dimension of capsules, the number of dynamic routing and  the hidden units of LSTM respectively .}
\label{tab.SepCapsNet param}
\setlength{\tabcolsep}{1.5mm}{
\begin{tabular}{c|cccc}
\hline
{\bf  Layer} & {\bf  CapsNums}& {\bf CapsDims}& {\bf Routing}&{\bf Hidden units}\cr
\hline
seq-caps layer 1 & {10}&{16}&{5}&{200}\cr
 \hline
seq-caps layer 2 &{5} & {10}&{5} & {128}\cr
\hline
\end{tabular}
}
\end{table}

\section{Results and Analysis}

%In order to illustrate the effectiveness of our proposed SECaps model,  we compare it against basic classical text classification methods  and two existing state-of-the-art charge prediction methods in three datasets. In addition, to prove the validity of our model in dealing with few-shot charge predictions, we run  a set of experiments with different frequencies for charge prediction. In particular, consider the influence of hyperparameters in our proposed SECaps model,  we run a set of experiments to evaluate our model by setting different parameters.

%In order to illustrate the effectiveness of our SECaps model,  we compare it with several state-of-the-art methods on three datasets. Then we run  a set of experiments with different frequency of charges to demonstrate the effectiveness of our model in dealing with few-shot charges.  In addition,  we conduct a series of ablation studies to evaluate the residual unit for the influence of our SECaps model. Besides, we consider the influence of hyperparameters in our SECaps model.

% In addition, to prove the validity of our model in dealing with few-shot charge predictions, we run  a set of experiments with different frequencies for charge prediction. In particular, consider the influence of hyperparameters in our proposed SECaps model,  we run a set of experiments to evaluate our model by setting different parameters.

\subsection{Performance Comparison}

\begin{table}
\centering
\caption{Charge prediction results of three datasets}
\label{tab:performance_comparison}
\setlength{\tabcolsep}{1.3mm}{
\begin{tabular}{c|cccc|cccc|cccc}
\hline
{\bf Datasets}&\multicolumn{4}{c|}{Criminal-S}&\multicolumn{4}{c|}{Criminal-M} &\multicolumn{4}{c}{Criminal-L}\cr\cline
{1-13}{\bf Metrics}&Acc.&MP&MR&MF&Acc.&MP&MR&MF&Acc.&MP&MR&MF\cr\hline
TFIDF+SVM&85.8&49.7&41.9&43.5&89.6&58.8&50.1&52.1&91.8&67.5&54.1&57.5\cr
CNN&91.9&50.5&44.9&46.1&93.5&57.6&48.1&50.5&93.9&66.0&50.3&54.7\cr
CNN-200&92.6&51.1&46.3&47.3&92.8&56.2&50.0&50.8&94.1&61.9&50.0&53.1\cr
LSTM&93.5&59.4&58.6&57.3&94.7&65.8&63.0&62.6&95.5&69.8&67.0&66.8\cr
LSTM-200&92.7&60.0&58.4&57.0&94.4&66.5&62.4&62.7&95.1&72.8&66.7&67.9\cr
\hline
Fact-Law Att.~\cite{Luo2017Learning}~&92.8&57.0&53.9&53.4&94.7&66.7&60.4&61.8&95.7&73.3&67.1&68.6\cr
Attribute-att.~\cite{hu2018few}~&93.4&66.7&69.2&64.9&94.4&68.3&69.2&67.1&95.8&75.8&73.7&73.1\cr
\hline
{\bf  SECaps Model} &{\bf 94.8}&{\bf 71.3}&{\bf 70.3}&{\bf 69.4}&{\bf 95.4}&{\bf 71.3}&{\bf 70.2}&{\bf 69.6}&{\bf 96.0}&{\bf 81.9}&{\bf 79.7}&{\bf79.5}\cr
\hline
\end{tabular}
}
\end{table}

Table \ref{tab:performance_comparison} shows the results of our model with baselines on three datasets. Overall, we find that the SECaps model outperforms all previous baselines with a significant margin on three datasets. More specifically, compared to the previous state-of-the-art in charge prediction \cite{hu2018few}, our model obtains $4.5\%$, $2.5\%$, and $6.4\%$ absolutely considerable improvements across three datasets respectively under MF, which demonstrates that the effectiveness of our SECaps model for charge prediction. This trend suggests that SECaps is capable of capturing  advanced semantic representation of legal texts  which are crucial for charge prediction.

%Based on this observation, we can infer the conclusion that our SECaps model generally beats the baselines and obtains the state-of-the-art performance.

%Ideally, our model learns useful information  by utilizing seq-caps layer, which brings global sequence information instead of partial information. Moreover, by applying the Multiple seq-caps layer and Attention mechanism, our model learns to directly capture the effective information and  guide benefit in making decisions. Consequently, they profit to  predict accurate predispositions, which lead to better performance.

We propose a novel layer, termed seq-caps, which considers sequence information and spatial information of legal texts simultaneously. Then our SECaps model employs multiple seq-caps layer to capture sequence information and advanced semantic representation which has a significant impact for charge prediction. In addition, our SECaps model introduces residual unit and designs an attention mechanism to capture significant auxiliary information of primary capsule for charge prediction. Consequently, our SECaps model obtains state-of-the-art performance on three datasets without any additional ancillary information.

\subsection{Few-shot Charges Comparison}

\begin{table}[htbp]
\centering
\caption{Macro F1 values of various charges on Criminal-S}
\label{tab.divide chargs}
\setlength{\tabcolsep}{1.8mm}{
\begin{tabular}{c|ccc}
\hline
{\bf Charge Type} & { Low-frequency}& {Medium-frequency}& {High-frequency}\cr\hline
{\bf Charge Number} & {49}  &
 {51}&{49}\cr\hline
 LSTM-200 &{32.6} & {55.0} & {83.3}\cr
 Attribute-att.~\cite{hu2018few}~ &{49.7} & {60.0} & {85.2}\cr\hline
 {\bf SECaps Model} &{\bf 53.8} & {\bf 65.5} & {\bf 89.0}\cr
\hline
\end{tabular}
}
\end{table}

Following Hu et al. \cite{hu2018few}, to further illustrate the effectiveness of the SECaps model on handling few-shot charges, we run a set of experiments to split charges with different frequency. We divide the charges into three parts according to the frequencies (low-frequency, medium-frequency and high-frequency). Low-frequency is defined as the charges appears less than 10 times (includes 10 times) on Criminal-S, high-frequency is defined as the charges appears more than 100 times (excepts 100 times) on Criminal-S and otherwise belongs to medium-frequency.

Table \ref{tab.divide chargs} shows the performance of our SECaps model with different frequency on Criminal-S, we report the low-frequency, the medium-frequency and the high-frequency results of MF. From the table we see that  the MF of low-frequency is $53.8\%$  which achieves more than $65\%$ improvements than LSTM-200 and obtains a considerable improvement by $4.1\%$  over the state-of-the-art baseline \cite{hu2018few}. Our SECaps model proposes a seq-caps layer, which can capture advanced semantic representation, and thus relieve the problem of insufficient features in the few-shot charge prediction. Specifically, the SECaps model has good power on  vector representation and time series representation ability, focal loss has a good performance in handling the problem of unbalanced classification, which can make up for lack of the unbalanced classification problem.

\subsection{Ablation Studies}

\begin{table}[!h]
\centering
\caption{
Ablation studies comparing our SECaps model with different
residual units on  Criminal-S.}
\label{tab:Ablation Study}
\setlength{\tabcolsep}{1.3mm}{
\begin{tabular}{c|cccc}
\hline
{\bf Models}&Acc.&MP&MR&MF\cr\hline
{\bf  SECaps Model} &{\bf 94.8}&{\bf 71.3}&{\bf 70.3}&{\bf 69.4}\cr
\hline
SECaps w/o Attention &94.7&67.7&68.1&66.4\cr
SEcaps with Added Unit   &94.6&66.1&65.3&64.0\cr
\hline
\end{tabular}
}
\end{table}

%\subsubsection{Comparison of Different  Residual Unit Methods}

In order to  evaluate the residual unit for the influence of our SECaps model, we conduct a series of ablation studies among various approaches. Table \ref{tab:Ablation Study} shows the results of various variant approaches.

${\bullet}$\quad \textbf{SECaps w/o Attention} which employs only two seq-caps layers to encode the primary capsules provides less performance when compared to our SECaps model. This signifies primary capsules aggregate higher-level capsules which only focuses on the most important legal case¡¯s information. It demonstrates that the residual unit is able to improve the generalization. Beside, it prove that the residual unit can provide auxiliary information for charge prediction.

${\bullet}$\quad \textbf{SEcaps with Added Unit} which employ simply added primary capsule unit instead of attention unit. As can be seen in Table \ref{tab:Ablation Study}, our SEcaps with added unit model even worse than  SECaps w/o Attention model. This phenomenon shows that the model brings some noise information if employ simply added primary capsules as the residual unit. Overall, attention unit can pay attention to the important information of primary capsules for charge prediction, which reinforce the SECaps model for capturing critical evidence.

\subsection{Impact of Hyperparameter}

\begin{figure}
\centering
\subfigure[The number of capsule.] { \label{fig:capsule_num}
\includegraphics[scale=0.25]{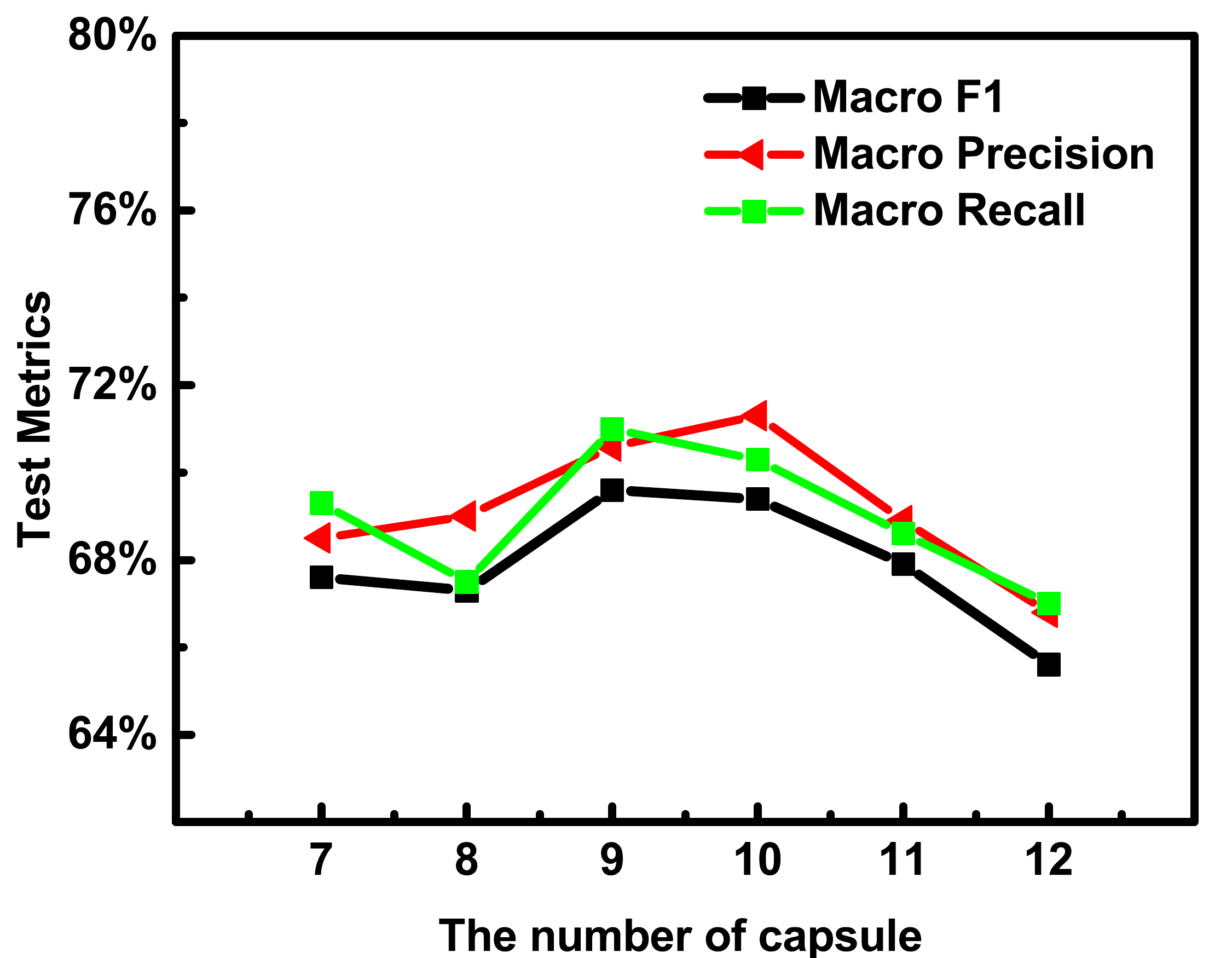}
}
\subfigure[The dimension of capsule.] { \label{fig:capsule_dim}
\includegraphics[scale=0.25]{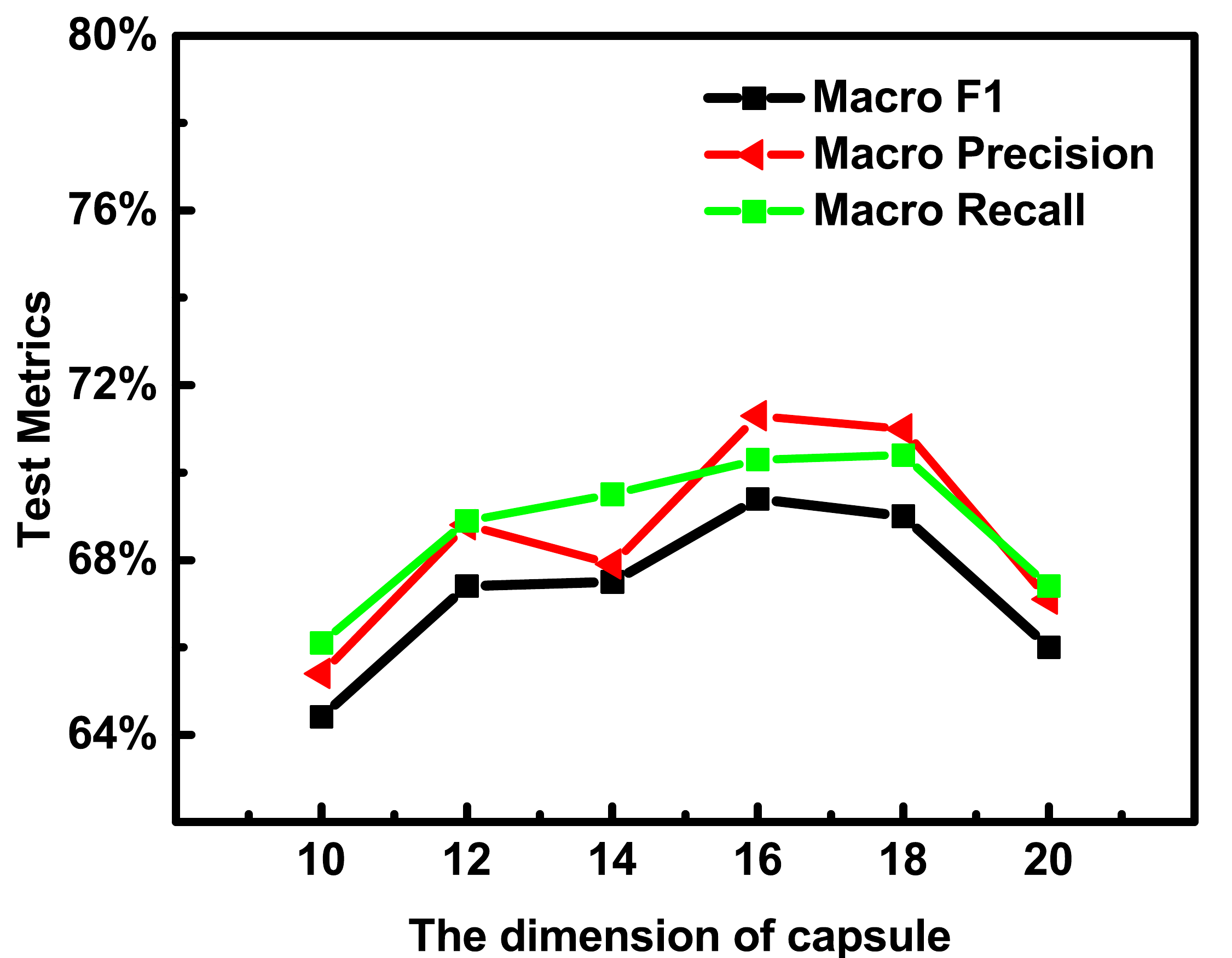}
}
\caption{Fig.\ref{fig:capsule_num} describes the relationship between MP,MR,MF and the number of capsules in seq-caps layer 1. % where the vertical axis denotes test
%metrics and the horizontal axis denotes the number of capsules. 
 Fig.\ref{fig:capsule_dim} describes the relationship between MP,MR,MF and the dimension of capsules in seq-caps layer 1.% where the vertical axis denotes test metrics and the horizontal axis denotes the dimension of capsules.
}
\label{fig:capsule_parm}
\end{figure}

%In the experiments, the number of capsules and the dimension of capsules are the most significant parameters in our SECaps model. 
%In order to evaluate the influence of two parameters for our SECaps model, we run a set of experiments by setting different values.
In this section,
%In particular, considering the influence of hyperparameter  in our proposed SECaps model, the two parameters that have the most influence on  the structure of our SECaps model are the number of the capsules and the dimension of capsules, we run a set of experiments to evaluate our model by setting different values.
we first study how the number of capsules affect the performance on Criminal-S. Our SECaps model are set the number of capsules from 7 to 12 in the seq-caps layer 1 and retained the rest of parameters unchanged. As shown in Figure \ref{fig:capsule_num}, we find that our SECaps model adds more capsules which can capture more vector representation. However, more capsules introduce noise which consequently decrease accuracy. %More specifically, our SECaps model increases the number of capsules which can constantly improve the vector representation of legal text and provide richer information. Nevertheless, it  brings a lot of redundant information, and increases the risk of overfitting, thus leading to the decline of the model's effectiveness. 
We set the parameter $CapsNums = 10$ in seq-caps layer 1 to balance the ability of representation of higher-level capsules.

We also study how the dimension of capsules affect the performance on Criminal-S. 
%Figure \ref{fig:capsule_dim} shows the comparison of ${\{10, 12 ,14, 16, 18, 20\}}$ dimensions  in the seq-caps layer 1.
  As shown in Figure \ref{fig:capsule_dim}
We find that our SECaps model obtains the state-of-the-art performance when dimension is set to 16.
The results indicate larger dimension's capsule is contributing to improve the performance. However, when the dimension of capsules is too large, the model aggregates more information from primary capsules and brings noise information which is helpless for charge prediction. Therefore, the dimension of capsules  should not be too large.

%Moreover, on purpose of illustrating the impact of number of capsules in our model, we run a set of experiments with different capsules. In particular, we compare the model with number of capsules from 7 to 12. For each time, we retain  the rest of the model unchanged, and report the F1, MP and MR  performance of our model on test dataset.

%Figure \ref{fig:capsule_num} shows the performance of our model in different capsules, We can see from the table that our model achieves better performance under Macro-F1 when the capsule number is equal to 9, while the model with the number of capsules is equal to 10  can obtain preferable effect under MP metric¡£Based on this observation, we can draw the conclusion that our proposed SECaps model generally achieves the state-of-the-art performance when the number of capsule is 10.

%Figure \ref{fig:capsule_dim} shows the performance of our model in different capsules, by seeing that the performance on test dataset stops increasing with number of capsule larger than 10, we therefore set the number of capsule to 10. It enables the model to directly optimize the  overall evaluation metrics, which yields a more effective and high-efficiency way to process tasks. Moreover, the dimensional of capsule is another significant factors which can influence the performance of our model.

\section{Related Works}
Automatic charge prediction plays an important role in the legal area and thus has received much attention. Researchers have proposed many methods for implementing automatic charge prediction. In this paper, these methods are classified into three categories: (1) traditional methods, (2) machine learning methods, and (3) deep neural network methods.

Traditional methods are usually mathematical or quantitative. Kort \cite{Kort1957Predicting} represents an attempt to apply quantitative methods to the prediction of human events. Nagel \cite{Nagel1964Applying} applys correlation analysis to case prediction. Keown \cite{keown1980mathematical} introduces mathematical (e.g., linear models and the scheme of nearest neighbors) models, which is used for legal prediction. These traditional methods have achieved some effects in certain scenarios, but they are restricted to a small datasets with few labels.

Researchers begin to use machine learning methods to handle charge prediction because of its success in many areas. This type of work usually focuses on extracting features from case facts and then using machine learning algorithms to make predictions. Liu et al. \cite{Liu2008Case,Liu2006Exploring} use K-Nearest Neighbor (KNN) method to classify criminal charges. Lin et al. \cite{Lin2014} fetch 21 legal factor labels for case classification. Mackaay et al. \cite{Ejan1974} extract N-grams features which creates by clustering semantically similar N-grams. Sulea et al. \cite{Sulea2017Exploring} propose a SVM-based system, which uses the case description, time span and ruling as features. However, these methods only extract shallow text features or manual tags, which are difficult to collect on larger datasets. Therefore, when the amount of the data is large, they will not perform well.

Recently, owning to success of deep neural network in the natural language processing (NLP), computer vision (CV) and speech fields, some works begin to apply the deep neural network to the charge prediction tasks and show a huge performance boost. Luo et al. \cite{Luo2017Learning} propose an hierarchical attentional network method, which predicts charges and extracts relevant articles jointly. Hu et al. \cite{hu2018few} propose an attention-based neural model by incorporating several discriminative legal attributes. The method proposed in this paper is classified into a deep neural network method.

%The most relevant to our work is the work of Hu et al. \cite{hu2018few}. Compared to this work, our work shares several common features with they: (1) we are both based on deep neural network methods  (2) we are both trying to solve the few-shot problem of charge prediction. Nevertheless, our work is different from this work in several features at least: (1)  the structure of the proposed model is different from them (2) the method to handle the few-shot problem in our model is different from them (3) we achieve the state-of-the-art performance for charge prediction as far as we known. We propose a Sequence Enhance Capsule model, dubbed as SECaps model, that can the prominent features and the semantic information of legal texts in a better way. Meanwhile, the SECaps model itself solves the few-shot through two strategies: (1) the seq-caps layer based on capsule can achieve very attractive results in few-shot charges datasets (2) the SECaps model introduces local loss, which can alleviate category imbalances to some extent.

Our work is also related to the task of text classification. Recently, various neural network (NN) architectures such as Convolutional Neural Networks (CNN) \cite{Kim2014Convolutional} and Recurrent Neural Networks (RNN) have been used for text classification. %Zhang et al. \cite{Zhang2015Character} offer an empirical exploration on the use of character-level convolutional networks for text classification.
Zhao et al. \cite{Zhao2018Investigating} explore capsule network with dynamic routing for text classification. From the perspective of using the capsule network, our work is related to Zhao et al. \cite{Zhao2018Investigating}. %But mainly differs in that we propose a new SECaps model based seq-caps layer, which can handle few-shot problem and achieve the state-of-the-art performance for charge prediction as far as we known.

%Another line of works that discussed zero-shot classification are also related to our work. Most works of zero-shot classification gain success in computer vision (CV). Jayaraman et al. \cite{Jayaraman2014Zero} introduce a random forest approach that explicitly explain the unreliability of attribute prediction. Akata et al. \cite{Akata2013Label} propose label embedding framework can transition smoothly from zero-shot learning to learning with large quantities of data. Lampert et al. \cite{Lampert2014Attribute} introduce attribute-based classification that the attribute classifiers can be pre-learned independently. Afterwards, new classes can be detected based on their attribute representation, without the need for a new training phase. Elhoseiny et al. \cite{Elhoseiny2014Write} make use of text description of the class label for zero-shot learning of object categories. Wu et al. \cite{Wu2014Zero} present a general framework for the zero-shot learning problem of performing high-level event detection. Zellers et al. \cite{Zellers2017Zero} model the visual and linguistic attributes of action verbs for large-scale zero-shot activity recognition. They also extend it to the few-shot scenarios.

Our model is loosely inspired from Hu et al. \cite{hu2018few}, they introduce several discriminative attributes of charges that provide additional information for few-shot charges. Compared with this line of works, although our work also handle the problem of few-shot charges, our work is different from their works, since (1) the strategy of our SECaps model in dealing with the problem of few-shot charges is different from the above (2) as far as we know, we are the first to introduce capsule network for charge prediction and achieve the state-of-the-art performance in charge prediction task.

\section{Conclusion}
In this paper, we focus on the few-shot problem of charge prediction according to the fact descriptions of criminal cases.
%This paper explores the problem of few-shot charge prediction according to the fact descriptions of criminal cases.
To alleviate the problem, we propose a Sequence Enhanced Capsule model for charge prediction. In particular, our SECaps model employs the seq-caps layer, which can capture characteristics of the sequence and abstract advanced semantic features simultaneously, and then combine with focal loss, which can handle the unbalanced problem of charges.
%This strategy contributes to the predicting for few-shot charges, bringing considerable chemical reaction for the task.
Experiments on the real-world datasets show that our SECaps model achieves $69.4\%$, $69.6\%$, $79.5\%$  Macro F1 on three datasets respectively, surpassing existing state-of-the-art methods by a considerable margin.

%In the future, we plan to explore neural network models for some cases that contain  multiple defendants and multiple charges in real-world. Additionally, we only utilize several simple attributes of charges, while there exist more complex essential conditions of charges. We plan to research on the possibility of applying transfer learning to predict more complex essential conditions of charges by our SECaps model.

\bibliographystyle{splncs04}
\bibliography{mybibliography}

\end{document}